# Modelling and Analysis of Walking Pattern for a Biped Robot

Aditya Gupta
The LNM Institute of Information and Technology
Jaipur, India
E-mail: adityagupta.y12@lnmiit.ac.in

Dr. Abhishek Shamra
Assistant Professor
The LNM Institute of Information and Technology
Jaipur, India
E-mail: abhisheksharma@lnmiit.ac.in

*Abstract*—This paper addresses the design and development of an autonomous biped robot using master and worker combination of controllers. In addition, the bot is wirelessly controllable. The work presented here explains the walking pattern, system control and actuator control techniques for 10 Degree of Freedom (DOF) biped humanoid. Bi-pedal robots have better mobility than conventional wheeled robots, but they tend to topple easily. In order to walk stably in various environments, such as on rough terrain, up and down slopes, or in regions containing obstacles, it is necessary, that robot should adapt to the ground conditions with a foot motion, as well as maintain its stability with a torso motion. It is desirable to select a walking pattern that requires small torque and velocity of the joint actuators. The work proposed a low cost solution using open source hardware-software and application. The work extends to develop and implement new algorithms by adding gyroscope and accelerometer to further the research in the field of biped robots.

## I. Introduction

Biped robots are anthropomorphic robots which can imitate the human gait. Research in bipedal robots is of keen interest these days, as they can be of great human aid. These are more responsive than wheeled robots, significantly while having motion in real-world scenarios such as uneven terrains, hill slopes and unknown obstacles [1] [2]. A number of correlated topics like robot kinematics, walking pattern generation, dynamics, gait simulation, mechanical design and static stability have been studied and researched upon. In recent years of humanoid research, many attempts have targeted on developing human gait inspired humanoids. This has resulted in development of robots like ASIMO, [3] also Samsung Electronics MAHRU [4]. Many such more efforts in this field were made, but still, the reconstruction of human gait is still not achieved.

Many researchers focused mainly on walking pattern generation. Since 1960s attempts have started taking place to build humanoid robots. One among the first working biped was developed by Kato et al. [5]. Walking pattern development using human gait kinematics was explored by Zerrugh et al. [6]. McGeer [7] styled a natural dynamic mode for walking by using inertia and gravity on a downhill slope. Methods for optimal energy intake cost function were proposed by Channon et al. [8] and Rostami et al. [9] through cohort of various gait functions. Further work on actuator's energy and power consumption was studied by Silva et al. [10]. The relationship between center of pressure (C.O.P) and zero moment point (Z.M.P) was correlated by Philippe at al. [11].

The need for human alike robots was felt in the situations where human life can be at stake or human existence is impossible. In addition, they can be used to imitate human body actions which can be of our assistance especially for elders and handicapped. It is used in the advancement of orthosis and prosthesis for human being. A few examples of its application include space exploration, application in biomedical sciences as replacement for damaged limbs. Sometimes it is required, fast and precise operations, for which anthropomorphic robots are better options than human as it eliminates the possibility of human error. Therefore, future research in this field is of great applications and importance.

Handling of several degree-of-freedoms (D.O.F) at the same time are the central events which we confront in the evolution of these automata, which is why stability is the primary event while developing walking pattern [12]. That is one of the major issues behind deficient fast and dynamic walking pattern, or anthropomorphic gaits, particularly on uneven terrain. Their tendency to topple easily makes their kinematics, even harder. The restriction with the size and weight of the hardware, which includes actuators and frame, restricts us to develop human sized robot. These factors directly affect the joints torque and velocity along with power consumption in doing so [13]. An optimal walking design needs a minimal requirement of the actuators i.e. minimum velocity and torque with less power consumption. By satisfying the above checks along with the stability, generation of an optimum walking pattern can be attained.

The work presented here for planning walking patterns comprises for even surface condition, stability restrictions and correct combination of actuators. In future, optimized walking patterns of the biped will be implemented to improve efficiency and robustness using inverted pendulum principle. This paper is organized as follows: The next section consists of literature review, in which there is detailed discussion on the concepts used, followed by modular description. After that experimental result are shown and explained, finally the paper is concluded followed by references in the end.

## II. LITRATURE RIVIEW

Moving further, prior knowledge of stability conditions along with the principle of inverted pendulum is to be defined for the development of walking pattern. Trailing these explanation of modules used is given, the hardware is chosen such that the effective cost of the robot is minimal and there is no compromises with the efficiency of results drawn.

### A. Definitions

I) Static Stability Definition: Following the definition given by McGheer and Frank, which states that an ideal robot is statistically stable if the horizontal projection of its center of gravity lies inside the support pattern which is on a horizontal surface. It consisted of a massless body and rigid trunk which could exert forces as desired. In static walking, the robot moves very slowly, hence its dynamics can be ignored. In addition, the projected center of gravity (PCOG) must be within the supporting area

II) Inverted Pendulum Principle: It states that, at the single support phase during human walking, the dynamical model of kinematics can be represented by a simple inverted pendulum as shown in fig(2), which connects the center of mass and the supporting base of the bipedal robot [14]. Stating above, ensures that the ground friction is large enough to prevent slipping. The principle evade the need of complex kinematics and dynamics calculations, hence is an easier and flexible way to plan motion for bi-pedals [15]. The dynamics equation for this principle can be easily derived using the law of energy conservation. The equation of motion of a simple inverted pendulum model can be written as follows:

$$T = mgl\theta - ml2\ddot{\theta} \text{ - (1)}$$

$$\text{Considering } F_z = mg \text{ ; } l\theta = Y_{mc}$$

$$\frac{T}{F_z} = Y_{mc} - \frac{l}{g}(\ddot{Y}_{mc}) \text{ - (2)}$$

$$Y_{ZMP} = Y_{mc} - \frac{l}{g}(\ddot{Y}_{mc}) \text{ - (3)}$$

Where T is torque acting on pendulum, m is mass, g is gravity constant, and Y is displacement. F is the force acting on the pendulum.

### B. Walking Patterns

While designing we need to keep in mind that in the walking pattern two forces acts simultaneously on bi-pedal robots, one transmitted by the contact and another transmitted without contact such as gravity, frictional forces etc. In addition, walking pattern is a periodic phenomenon [16] [17].

A coherent design was considered with each leg consisting of five degree of freedom (DOF). They are as follows: a hip joint, a thigh joint, a knee joint, a shin joint and a foot joint as shown in fig(1). The complete walking cycle consists of two phases: single support phase and double support phase. In sagittal plane movement, during single support phase as well as double support phase the COG must lie between two[18]. During single phase, one leg is stationary on ground while other leg is in air. In double support phase both feet are in direct contact with level surface. It initiates when heel of the front foot drops on ground, and, ends when toe of the other foot parts away from ground. The unilaterality of the foot-ground contact is deciding factor in locomotion of biped [19] [20]. In fact, the pressure force is focused on foot, away from the ground.

Moving on to gait designing, intermediate joint angles are decided such that the center of mass (COM) is always on desired location and lift of swinging leg is achieved with efficiency. Walking on non-ideal surface with obstacles, it is necessary for biped to raise the swinging foot high enough to avoid obstacles [21] [22]. The key parameter to keep in mind while modelling biped for uneven surfaces is stability. The bot should also be stable enough to sustain external forces such as air pressure, during the single support phase. Much of the current research on humanoid robot is done to encounter real life situations as mentioned above.

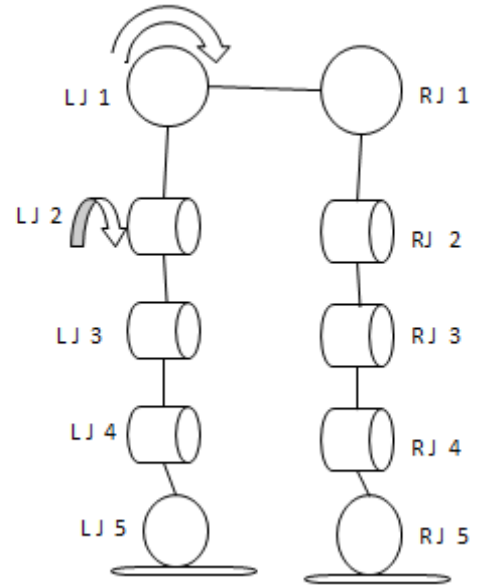

Fig 1: Model of Biped Robot

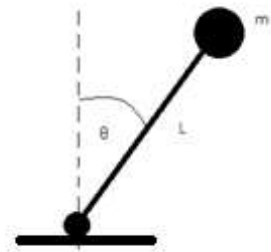

Fig 2: Inverted Pendulum Model

## C. Modular Descriptions

I) Arduino Uno: It is the brain of our system, which contains the code and accepts data from transmitter. It sends commands to servo controller to make them move in desired way. It consists of ATMEGA328P microcontroller and 6 analog input/output pins, from which various sensors can be attached, to improve the walking ability of biped. This is an open source platform which makes it even more cost effective.

II) Master Slave Communication: In this model of communication one device has unidirectional control over one or more devices. In this case, the controlling unit acts as master while, the module connected to servos acts as slave. The setup configuration is shown in fig (3).

III) Wireless Communication Module: "PS2 Wireless Controller" was used to control the actions of biped robot. The receiver was connected to microcontroller to receive the user commands, and is supported by almost every microcontroller in the market. It is very economical and effective to avoid human intervention in controlling such bots. It has wide application in the areas where robots can be used for rescue missions and search operations.

IV) Pulse Width Modulation: Encoding technique, for controlling analog circuits. It uses a rectangular pulse wave whose pulse width is modulated resulting in the variation of the average value of the waveform. In this model, servo controller module generate PWM signal which then gets transmitted to servos and controls them. PWM signal waveform is shown in fig(4). Pulse width modulation is fine technique to send digital signal in analog manner.

V) Servo Motors: Servos are actuators, controlled by pulse width modulated signals. The angle of its rotation is determined by the on time of the pulse, in one time period which is 20 ms. Servo has three wires, red wire for power, black wire for ground and white wire to receive PWM signals. It is recommended to use different power sources for servos from controllers as they are electrically noisy elements. In this model we used ten MG-996 servos with 10 kilogram torque capacity, with operating voltage of 4.8-7.2 V.

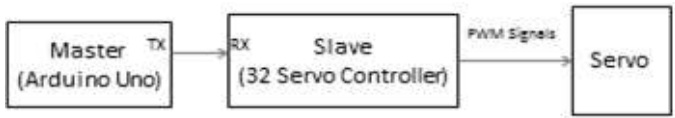

Fig 3: Block Diagram for Communication

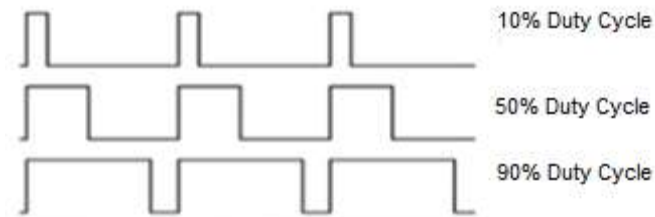

Fig 4: PWM Signals

## III. EXPERIMENTAL RESULTS

As shown in the fig (5), the model consists of 10 degrees of freedom (D.O.F) with five in each leg. Joint pairs (RJ4 and RJ5) and (LJ4 and LJ5) are attached such that foot can perform planar flexion, eversion and inversion. Moreover, the joint pairs (RJ1 and RJ2) and (LJ1 and LJ2) are such that hip can perform flexion, extension, abduction and adduction.

For the forward mode of operation, initially, in *stage I* the robot is in the double support phase with both legs having equal joint angles, and in a state of stable standing. The center of mass lies exactly in between of bi-pedal robot model, and therefore, it's in the static stability mode.

Elble et al.[17] have found that center of mass shifts in lateral and anterior direction while moving forward, therefore, weight from swing leg is shifted in the direction of the stance leg, this is due to the fact that the torque in stance leg's hip increases. Swing leg is the one in air and stance leg is one on the ground. With this *stage II* was designed, in which single support phase is achieved by rotating (LJ2, LJ3 and LJ4) in anti-clockwise direction, such that left leg attain appropriate flexion and (RJ5 and RJ1) are rotated in a clockwise direction to abduct so as to maintain stability.

In *stage III* double support phase is achieved by left leg again touching the ground. During this transaction the center of mass shifts to the center of the body. While achieving this stage (LJ2, LJ3 and LJ4) are moved in the clockwise direction to straighten up the left leg and (LJ5) is moved in anti-clockwise direction to obtain contact phase. In the right leg (RJ1 and RJ5) obtain zero displacement and (RJ2, RJ3 and RJ4) move in the clockwise direction for maximum extension. Foot of the right leg is in propulsive phase giving the boat static stability and correct positioning of the center of mass.

In *stage IV* the right leg is in the air and left leg acts as a single support for a biped. The center of mass shifts towards left leg, as it abducts itself and right leg achieve appropriate flexion. During this, (LJ5 and LJ1) are moved in anti-clockwise direction, as well as (LJ2, LJ3 and LJ4) are moved in the clockwise direction to balance center of mass. And, in the right leg (RJ2) is rotated in anti-clockwise direction and (RJ3 and RJ4) in clockwise direction. At the end of the phase, right leg is ahead of left leg as shown in fig (5).

In *stage V* right leg lands on the ground and biped achieve double support phase. While commencing this stage, (R2, R3 and R4) are rotated in the clockwise direction and (LJ5 and LJ1) are moved in clockwise direction to obtain stable structure with proper location of center of mass. Subsequently, *stage 1* is reiterated to establish in its initial position, i.e. *stage1*.

As shown in table (1A and 1B), the initial values of PWM are as in *Stage 1* where all joints are such that bot stands erectly. The changes done in PWM values are done such that they acquire required angle. The values sometimes varied as the servo did not responded ideally. We could give signals ranging from 800 to 2400 to servos either of them represents clockwise or anticlockwise extremes depending on their orientation in model. The values in the table will change accordingly if the values in *stage 1* (initial values) are changed. Table 1A contains the value for right leg while table 1B contains the value for left leg. For continues walking *stage 1* is repeated after *stage 5*.

|  | Stage 1 | Stage 2 | Stage 3 | Stage 4 | Stage 5 |
|---|---|---|---|---|---|
| R J 1 | 870 | 891 | 891 | 870 | 870 |
| R J 2 | 1152 | 1043 | 1043 | 1152 | 826 |
| R J 3 | 957 | 1043 | 1043 | 957 | 957 |
| R J 4 | 957 | 935 | 935 | 957 | 1109 |
| R J 5 | 1696 | 1500 | 1500 | 1500 | 1522 |

Table 1A: PWM values for right leg joint.

|  | Stage 1 | Stage 2 | Stage 3 | Stage 4 | Stage 5 |
|---|---|---|---|---|---|
| L J 1 | 2152 | 2152 | 2152 | 2152 | 2152 |
| L J 2 | 1043 | 1087 | 826 | 1043 | 1043 |
| L J 3 | 957 | 891 | 891 | 957 | 957 |
| L J 4 | 1935 | 1913 | 1913 | 1935 | 1935 |
| L J 5 | 1761 | 2000 | 1783 | 1826 | 2000 |

Table 1B: PWM values for left leg joint.

Table 1A and 1B contains the mutually synchronized PWM values of all the servos for forward motion of the bipedal robot.

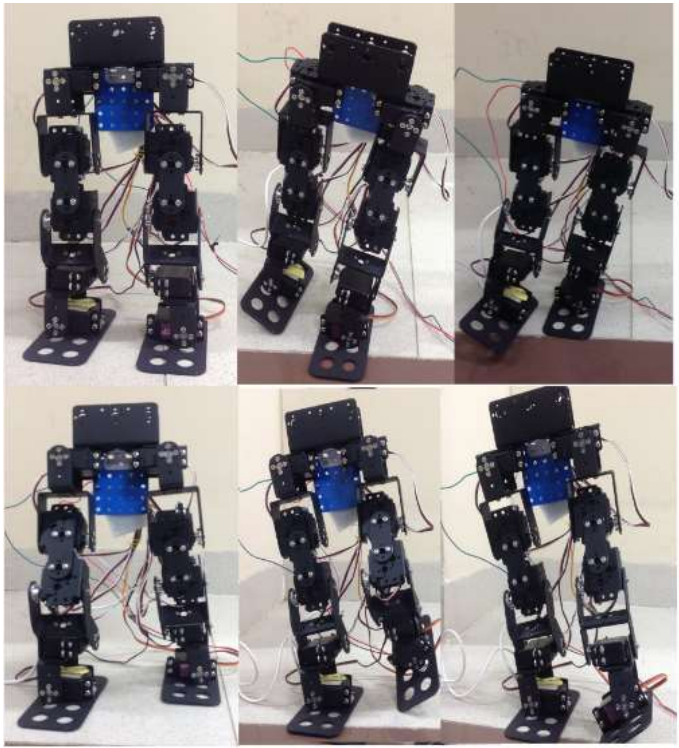

Fig 5: Forward Mode Walking Stages.

Figure 5 shows the real time walking of the bi-pedal robot. In first row left legs motion is shown and in the second row right leg motion is shown. As shown, initially robot is in in stable standing mode. Also, while performing the experiment the surface was even and external forces and obstacles were removed.

IV. CONCLUSION

Owing to the increasing importance of humanoid robot further research is required. Especially in places where human presence can be hazardous for example in rescue missions. The use of humanoid robots for the physically disabled can be of great importance and is yet to be researched upon. Handling robot through remote control adds on to its applications in dangerous and remote locations. In this work, we focused on analyzing and improving the movement of such a machine. The walking pattern was motivated from the movement of an inverted pendulum. The idea was realized by the use of Atmega 328 microcontroller and commonly available hardware for actuators, making this work highly viable in real life scenarios. The results demonstrated that our robot could perform walking action in a highly stable manner. We believe that further improvements can be done with the actions by adding voice recognition and artificial intelligence. In the future, we would be working to improve the pattern by feedback correction from sensors.